# Non-rigid Reconstruction with a Single Moving RGB-D Camera

Shafeeq Elanattil[1,2], Peyman Moghadam[1,2], Sridha Sridharan[2], Clinton Fookes[2], Mark Cox[1]

[1]Autonomous Systems Laboratory, CSIRO Data61, Brisbane, Australia
[2]Queensland University of Technology, Brisbane, Australia
{shafeeq.elanattil, peyman.moghadam, mark.cox}@data61.csiro.au
{s.sridharan, c.fookes}@qut.edu.au

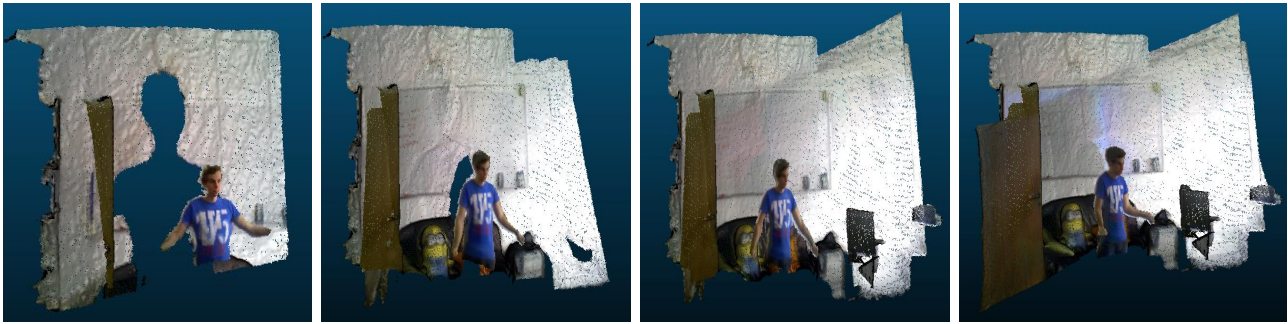

Fig. 1: Scene reconstruction results at different time instances using the proposed approach. Our method provides a complete reconstruction of rigid and non-rigid objects in the scene

*Abstract*—We present a novel non-rigid reconstruction method using a moving RGB-D camera. Current approaches use only non-rigid part of the scene and completely ignore the rigid background. Non-rigid parts often lack sufficient geometric and photometric information for tracking large frame-to-frame motion. Our approach uses camera pose estimated from the rigid background for foreground tracking. This enables robust foreground tracking in situations where large frame-to-frame motion occurs. Moreover, we are proposing a multi-scale deformation graph which improves non-rigid tracking without compromising the quality of the reconstruction. We are also contributing a synthetic dataset which is made publically available for evaluating non-rigid reconstruction methods. The dataset provides frame-by-frame ground truth geometry of the scene, the camera trajectory, and masks for background foreground. Experimental results show that our approach is more robust in handling larger frame-to-frame motions and provides better reconstruction compared to state-of-the-art approaches.

## I. Introduction

The advent of inexpensive RGB-D sensors, combined with the availability of GPUs for high-performance computing has resulted in great progress for dense reconstruction in real-time. While a large number of solutions exist for reconstructing static scenes [1], [2], [3], the more general scenario, where the objects are dynamic and undergo non-rigid deformation, is still a challenge to be solved [4], [5], [6], [7].

The large solution space and inherent ambiguity of deformation makes this problem intractable. External constraints are used for alleviating these issues like carefully designed capture environments [8], multiple video cameras [9], template-based approaches [10] and high-quality lighting equipment [11]. DynamicFusion [4] is the pioneering work, which incrementally reconstructs a non-rigid 3D scene from the single RGB-D stream. This method bypasses the need for a template for non-rigid reconstruction and can operate in real-time for augmented-reality or virtual-reality applications. More recent works extended DynamicFusion by reducing drift using image features [5], adding shading information for better tracking [6] and introducing Signed Distance Function (SDF) based flow vectors for addressing problems due to topology changes [7].

State-of-the-art approaches [4], [5], [6], [7] use only the non-rigid part of the scene and completely ignores the rigid background. Non-rigid part of the scene often lacks enough geometric and photometric features for tracking in larger baseline situations. For addressing aforementioned limitation, we propose to use camera pose estimated from the rigid background for tracking the non-rigid object. Our approach uses a segmentation method to separate non-rigid and rigid parts of the scene. We also propose a multiscale deformation graph, which helps to track a wider range of deformations without compromising reconstruction quality. Figure 1 shows an example of our reconstruction results at different frames. We have developed a synthetic dataset for evaluating RGB-D based non-rigid reconstruction methods. The dataset contains the per frame ground-truth of output geometry and camera trajectory. The key contributions of our approach can be summarized as follows:

- The camera pose estimated from the background enables our approach to achieve more robustness in handling larger frame-to-frame motions.
- The multiscale deformation graph enables tracking a wider range of deformations without compromising the quality of reconstruction.
- Our synthetic dataset provides frame-by-frame ground truth geometry of the scene, the camera trajectory and masks for background and fore-ground. This enables quantitative evaluation of per-frame reconstruction among non-rigid reconstruction approaches.

## II. RELATED WORK

The non-rigid 3D reconstruction is a widely investigated topic in computer vision. The large parameter space for modeling non-rigidity makes this problem challenging. There are mainly three categories of approaches. The first category uses priors as a template or multiple cameras for making the problem tractable. The second category incrementally reconstructs the scene without any template (template-free). Third category directly process 3D points from the sensor for rendering the geometry.

### A. Multi-view and Template based approaches

Vlasic *et al.* [12] used a skeleton based human body template for modeling articulated motion in a multi-view setting. The parameter space is reduced to the joint angles between the skeleton bones. However, this limits the range of deformations that can be modeled. Zollhöfer *et al.* [10] used a deformation graph for tracking non-rigid motions in the scene based on a template . They utilised Iterative Closest Point (ICP) like correspondences and a local rigidity assumption for tracking. Although these techniques achieve accurate non-rigid tracking for a wide variety of motions, they require an initial template geometry as a prior. For this purpose, the non-rigid object has to be still during template generation, which cannot be guaranteed in general situations. Recently, Dou *et al.* [9] demonstrated a non-rigid reconstruction system using 24 cameras and multiple GPUs, which is a setup not available to a general user.

### B. Template-free approaches

Template-free approaches incrementally reconstruct the scene by tracking deformations simultaneously. This kind of method is desirable for single sensor systems. Dou *et al.* used a non-rigid bundle adjustment for reconstruction of a non-rigid subject [13], but it takes 9-10 hours for optimization. DynamicFusion was the first approach to simultaneously reconstruct and track a non-rigid scene in real-time [1]. VolumeDeform extended this work by using SIFT features across all images to reduce drift [5]. Both of the approaches provide compelling results in relatively controlled motions.

Guo *et al.* [6] improved tracking by using surface albedo, which is estimated using lighting coefficients under Lambertian surface assumption. However Lambertian surface assumption only holds in a controlled lighting scenarios. Slavcheva *et al.* [7] used SDF based directional field for tracking deformations. This approach can track greater range of motions and solve problems due to topological changes up to some extent. However, the SDF based optimization significantly increases the computational overhead. Therefore, this approach can only produces coarse reconstructions in real-time.

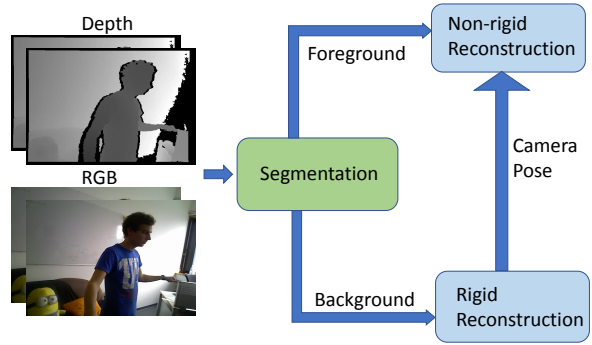

Fig. 2: Block diagram of the proposed method.

### C. Point based approaches

3D points are a natural primitive for geometry acquisition in most devices. Wand *et al.* [14] proposed a point based surface representation for temporal fusion to build a template of dynamic objects. Kuster *et al.* [15] used Moving Least Square (MLS) based surface approximation for modelling dynamic object. Similar to our approach, they also reconstructing rigid and non-rigid parts of the scene. However, the unstructured nature of the points makes these approaches computationally expensive and not suitable for real-time applications, whereas the structured data-structure (TSDF) in our approach achieves real-time performance by using Graphical Processing Units (GPU) threads.

## III. OVERVIEW

Figure 2 shows the high-level block diagram of the proposed approach. Our segmentation method separates RGB-D data into foreground and background. The non-rigid and rigid reconstruction modules process foreground and background separately. The camera pose estimated by rigid reconstruction module is used to improve non-rigid tracking. Our method uses a multi-scale deformation graph for non-rigid tracking. Following subsections explain each module of the proposed approach.

### A. Segmentation

Figure 3 shows the pipeline of the proposed segmentation. This method operates in two steps. In the first step, the depth image $D$ is segmented using Connected Component Labeling (CCL) algorithm [16]. CCL segments by a Euclidean comparison function $C$. Let the Euclidean distance

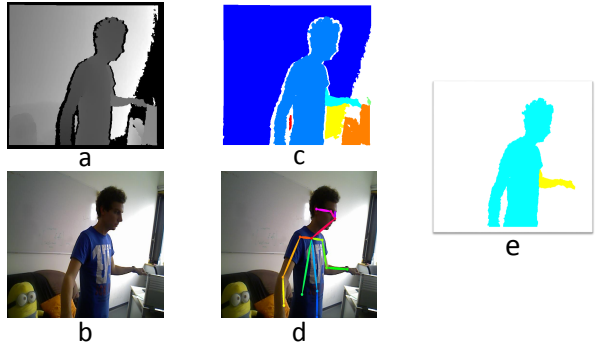

Fig. 3: Pipeline of our segmentation method. Input (a) depth and (b) RGB images. (c) Labelled depth image after connected component labelling. (d) Skeleton detected from the RGB image. (e) Final segmented foreground.

between $P(x_1, y_1)$ and $P(x_2, y_2)$ be $dist$, then the comparison function is

$$C(P(x_1,y_1), P(x_2,y_2)) = \begin{cases} \text{true}, & \text{if } dist < \theta_{th} \\ \text{false}, & \text{otherwise.} \end{cases} \quad (1)$$

Here $P(x, y)$ is the 3D point corresponding to the location $x,y$ of the depth image. We took $\theta_{th}$ equal to 4mm in our experiments.

In the second step, we use a Convolutional Neural Network (CNN) based human pose detection algorithm [17]. This algorithm detects human pose and draws the skeleton in the image. Then we classify segments (from CCL) containing skeleton bones as foreground and rest as background. The motivation of this segmentation module is to separate background and foreground for the rest of the pipeline. This approach works well for simple situations where a human does activities while standing on the ground. Note that because of our focus is on reconstructing human subjects, we use a human segmentation method. Our approach can easily be extended to other non-rigid subjects simply by changing segmentation for respective non-rigid objects.

*B. Non-rigid Reconstruction*

Similar to DynamicFusion [4] our approach operates in a frame-to-frame manner. For each frame, the canonical model of the scene is developed in SDF using weighted average scheme of Curless and Levoy [18]. A warp field $\mathbf{W}$ is used for integrating data from each frame. $\mathbf{W}$ gives a mapping from canonical world coordinates to live camera coordinates. The warp field is also used for making live reconstruction. Our warp field has two components, 1) Rigid component 2) Non-rigid component. The warp function $\mathbf{W}$, of a point $\mathbf{x_c}$ in canonical coordinate is of the form as shown below,

$$\mathbf{W}(\mathbf{x_c}) = \mathbf{T_{rigid}} \mathbf{W_{non-rigid}}(\mathbf{x_c}), \quad (2)$$

where $\mathbf{T_{rigid}}$ is the rigid and $\mathbf{W_{non-rigid}}$ is the non-rigid components warp function. $\mathbf{T_{rigid}}$ is $4 \times 4$ rigid transformation matrix and $\mathbf{W_{non-rigid}}$ is modelled by deformation graph [19]. Each graph node stores the rigid transformation matrices $\{\mathbf{T_i}\}$. For a point $\mathbf{x}$, transformed by warp field $\mathbf{W_{non-rigid}}$ as shown below,

$$\mathbf{W_{non-rigid}}(\mathbf{x}) = \sum_{\mathbf{p_i} \in \mathcal{N}(\mathbf{x})} w(\mathbf{p_i}, \mathbf{x}) \mathbf{T_i} [\mathbf{x^T} 1]^T, \quad (3)$$

where $w(\mathbf{p_i}, \mathbf{x})$ is the influence weight of point $\mathbf{x}$ to node $i$, $\mathcal{N}(\mathbf{x})$ is the set of nearest graph nodes from $\mathbf{x}$ and $\mathbf{p_i}$ is the position of the $i^{th}$ graph node. Similarly normal $\mathbf{n}$ at the point $\mathbf{x}$ is transformed by warp field $\mathbf{W_{non-rigid}}$,

$$\mathbf{W_{non-rigid}}(\mathbf{n}) = \sum_{\mathbf{p_i} \in \mathcal{N}(\mathbf{x})} w(\mathbf{p_i}, \mathbf{x}) \mathbf{T_i} [\mathbf{n}^T 0]^T. \quad (4)$$

*1) Estimation of Rigid component ($\mathbf{T_{rigid}}$):* Current approaches [4], [5], [7], [6] use a rigid Iterative Closest Point (ICP) step for estimating $\mathbf{T_{rigid}}$. They just use foreground and completely ignore the background. In most cases the foreground is smaller compared to the background. Therefore tracking often fails during large frame-to-frame motion. For overcoming this limitation, we use the camera pose from the background $\mathbf{T_{bg}}$ for estimating $\mathbf{T_{rigid}}$. This is based on an observation that since both $\mathbf{T_{rigid}}$ and $\mathbf{T_{bg}}$ depend upon camera motion, $\mathbf{T_{bg}}$ becomes a better initializer for estimating $\mathbf{T_{rigid}}$.

For computing $\mathbf{T_{rigid}^t}$ at a time $t$, we take incremental background transformation $\mathbf{T_{bg}^{inc}}$ (computation of incremental background transformation $\mathbf{T_{bg}^{inc}}$ is described in Section III-C) from rigid reconstruction module and update the previous $\mathbf{T_{rigid}^{t-1}}$ as,

$$\mathbf{T_{rigid}^*} = \mathbf{T_{bg}^{inc}} \mathbf{T_{rigid}^{t-1}}. \quad (5)$$

Then we update warp function $\mathbf{W_{t-1}}$ from last frame as below

$$\mathbf{W_t^*} = \mathbf{T_{rigid}^*} \mathbf{W_{t-1}}, \quad (6)$$

based on intermediate warp function $\mathbf{W_t^*}$ we transform foreground vertexes $\mathbf{v_{fg}}$ and normals $\mathbf{n_{fg}}$ using the Equations 2, 3, 4 and 6.

$$\mathbf{v_{fg}^*} = \mathbf{W_t^*}(\mathbf{v_{fg}}) \quad (7)$$

$$\mathbf{n_{fg}^*} = \mathbf{W_t^*}(\mathbf{n_{fg}}) \quad (8)$$

Now an ICP step is performed for finding incremental transformation $\mathbf{T_{rigid}^{inc}}$ minimizing point normal distance between transformed vertexes $\mathbf{v_{fg}^*}$ and vertexes from the current frame ($\mathbf{v_{fg}^s}$). This achieved by minimizing the following objective function.

$$E_{icp}(\mathbf{T_{rigid}^{inc}}) = \sum_{i,j} ||(\mathbf{v_{fg}^*}(i) - \mathbf{v_{fg}^s}(j)).\mathbf{n_{fg}^*}(i)||^2, \quad (9)$$

where $i,j$ are indexes of the the correspondence between $\mathbf{v_{fg}^*}$ and $\mathbf{v_{fg}^s}$. Correspondences are estimated using projective data association. After this step $\mathbf{T_{rigid}^t}$ is estimated as,

$$\mathbf{T_{rigid}^t} = \mathbf{T_{rigid}^{inc}} \mathbf{T_{rigid}^*}. \quad (10)$$

By this we using full available information of depth image for tracking, which makes foreground tracking more robust during large frame-to-frame motion.

*2) Non-rigid tracking (Estimation of $\mathbf{W_{non-rigid}}$):* We use deformation graphs for non-rigid tracking, in which nodes are created by subsampling the canonical model and edges are formed by connecting nearest neighbors. State-of-the-art approaches [4] [6] use single deformation graph of a specific sampling density. We observe two trade-offs due to the sparsity of sampling density. 1) Even-though a sparser graph can track more range of articulated motion, it reduces overall reconstruction quality. 2) A denser graph can reconstruct finer deformations however, it is more susceptible to converge into local minima. Therefore, for improving tracking without compromising the quality of reconstruction, we propose a multiscale deformation graph for non-rigid tracking. Our multiscale approach consists of two graphs a dense graph $G_d$ and a sparse graph $G_s$. For each frame we first align the sparse graph with the incoming frame. Then parameters of sparse graph $G_s$ is transformed to dense graph using linear blending. Finally, we align dense graph with the incoming frame. Both graphs use a same optimization objective for aligning with data.

The deformation parameters of the graph $\{\mathbf{T_i}\}$ is estimated by minimizing the below objective function.

$$E_{non-rigid}(\{\mathbf{T_i}\}) = \alpha_{data} E_{data} + \alpha_{reg} E_{reg} \quad (11)$$

**Data term** $E_{data}$ represents the error of data fitting from the correspondence points. We are using projective data association for getting correspondence points. Data term $E_{data}$ is point to plane error metric same as Equation 9.
**Regularization term $\mathbf{E_{reg}}$** is the local as-rigid-as possible constraint imposed on neighbouring graph nodes. This term is important for driving invisible regions to move with the observed regions by generating smooth deformations.

$$E_{reg} = \sum_i \sum_{j \in \mathcal{N}(i)} ||\mathbf{T_i p_i} - \mathbf{T_j p_j}||^2 \quad (12)$$

where $\mathbf{T_i}$, $\mathbf{T_j}$ are transformation associated with each graph nodes, $\mathbf{p_i}$, $\mathbf{p_j}$ are the positions of the graph nodes and $\mathcal{N}(i)$ represents nearest neighbour set of $i^th$ node in the graph. The objective function (Equation 11) represents a non-linear least square problem and solved by a GPU based Gauss-Newton solver.

*3) Canonical Model Update:* After the optimization step, we integrate current depth information into a reference volume using a weighted average scheme. Based on the newly added regions we update both the dense and sparse graphs as explained in [4].

### C. Rigid Reconstruction

For each input frame, we track camera pose $\mathbf{T_{bg}}$ by performing an ICP step between current point cloud $\mathbf{V_{bg}^s}$ with transformed point cloud from the previous frame, obtained by projecting the canonical model using the estimated pose $\mathbf{T_{bg}^{t-1}}$ of the previous frame. This is similar to other RGB-D based rigid reconstruction approaches [1].

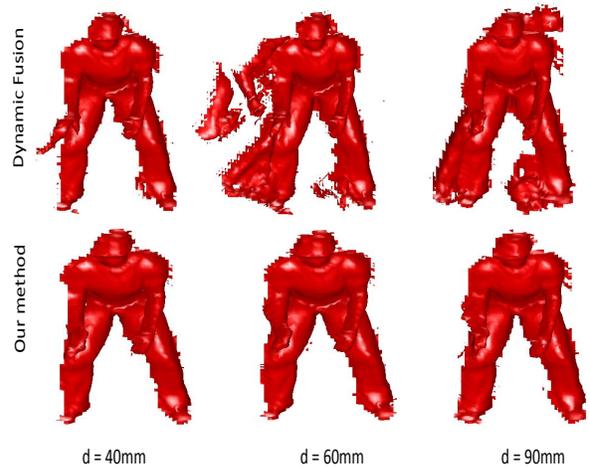

Fig. 4: Snap-shots of reconstruction from DynamicFusion [4] and our approach are shown. Here $d$ is the frame-to-frame distance in our saw-tooth camera trajectory (refer section V-A).

### D. Implementation Details

Our approach has two main contributions over and above the implementation reported in [4]. (i) Segmentation module and (ii) multi-scale deformation graphs. The segmentation module consists of a CNN based skeleton detector and a CCL module. Our single GPU based system takes 200ms and 3ms per frame for CNN and CCL module respectively. Because of a highly parallelizable architecture of CNN module, the processing time can be further reduced to around ~30ms using the techniques proposed in [20]. The multi-scale deformation graph is developed in the same way as in [4]. The main difference is, instead of four non-rigid ICP iterations, our approach uses two non-rigid ICP iterations with each coarse and fine graphs.

## IV. SYNTHETIC DATA

There are only a few public data sets available for evaluating RGB-D based non-rigid reconstruction approaches. The dataset published with VolumeDeform [5] grants canonical and live reconstruction of their approach at every 100th frame. This can only be used for comparing with their approach. The dataset published with KillingFusion [7] has the canonical ground truth. This can be used for comparing canonical reconstruction. Both these approaches not providing any ground truth for comparing live reconstruction.

For evaluating live reconstruction, reconstruction should be in world coordinates. Otherwise camera pose is needed for transforming live reconstruction to world coordinates. All previous approaches [4], [5], [6], [7] do not estimate camera

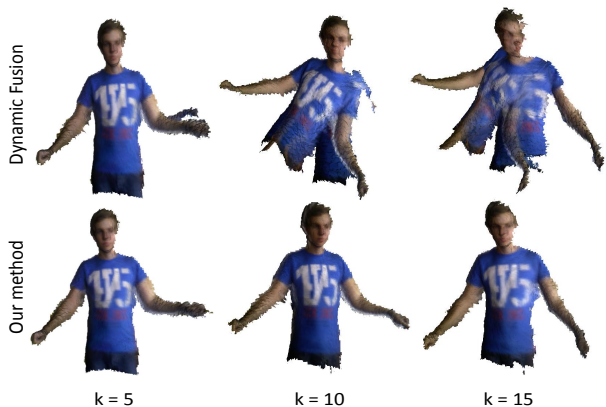

Fig. 5: Snap-shots of reconstruction from the "Upperbody" data from VolumeDeform [5]. Upper row shows reconstruction from DynamicFusion [4] and lower row shows reconstruction from our method. Here $k$ is the number of frames skipped for simulating the motion of $k\times$ speed (refer section V-A).

|  | d=40mm | | d=60mm | | d=90mm | |
|---|---|---|---|---|---|---|
|  | mean | std. | mean | std. | mean | std. |
| DynamicFusion | 11.96 | 6.4 | 17.64 | 11.69 | 18.73 | 11.61 |
| Our method | **10.44** | **5.6** | **11.58** | **5.54** | **11.65** | **5.49** |

TABLE I: Comparison of the squared distance error from ground truth by using DynamicFusion [4] and our method. Here $d$ is the frame-to-frame distance in our saw-tooth camera trajectory (refer section V-A). The unit for error is mm.

pose, therefore these approaches cannot be compared with live ground truth. Since our approach estimates camera pose, we are able to compare the live reconstruction with live ground truth. Using this as motivation, we developed a synthetic dataset for complete evaluation of single RGB-D based non-rigid reconstruction approaches. Our dataset consists of

1) Complete scene geometry at first frame for evaluating canonical reconstruction.
2) Live scene geometry at each frame in world coordinates.
3) Ground truth of camera trajectory.
4) Ground truth of foreground mask.

For making synthetic data, at first, we make a human body model using MakeHuman package[1]. MakeHuman is an open source tool for making 3D human characters. Then we export this character to Blender[2]. Blender is an open-source 3D computer graphics toolset where the 3D characters can be deformed programmatically. Then we use Carnegie Mellon University (CMU) motion capture dataset[3] for animating the 3D model. This enables us to simulate a wide variety of human movements for experimenting. Furthermore, Blender enables placing a moving camera around the scene with which RGB and depth images are synthesized. This enables us to get the ground truth camera pose for evaluation. We have made our database publicly available to enable researchers to replicate our results and contribute to further advances in the area[4].

## V. EXPERIMENTS

Reconstruction results of some human motion sequences are are given in our supplementary video. In this section, we describe the qualitative and quantitative evaluation of our reconstruction framework.

### A. Robust tracking in large frame-to-frame motion

At first, we evaluate the advantage of using background transformation for tracking foreground. In order to test this innovation, we collect data using a saw-tooth camera trajectory over a non-rigid synthetic scene. The saw-tooth trajectory is created by moving to and fro a distance $d$ between adjacent frames in a circular trajectory. Then reconstruction is made using our approach and DynamicFusion [4]. DynamicFusion and all the other state-of-the-art approaches [5], [6], [7] only use foreground for tracking. We found both approaches work well until $d$ equal to 20mm and DynamicFusion gives erroneous reconstruction for all tested values of $d$ equal to 40mm, 60mm and 90mm respectively. Snapshots of both reconstructions are shown in the Figure 4. Table I tabulates the average and standard deviation of errors from both approaches. Our approach has less error in all three cases and works well for frame-to-frame distance $d$ of up-to 90mm. Note that we run this experiment by using ground truth foreground masks for to ensure that the segmentation errors have no effect on the final reconstruction.

We also evaluate the tracking stability in real data. We test with Upperbody data from VolumeDeform [5] dataset. We reconstruct the scene by using every $k^{\text{th}}$ input frame (k = 3,5,10,15). This simulates motion of $k\times$ speed. We find both approaches work well when $k$ less than 5 and DynamicFusion gives erroneous reconstruction for the tested values of k equal to 5, 10 and 15. Snapshots of both reconstructions are shown in the Figure 5. Table II tabulates the average and standard deviation of errors from both approaches in these cases. Our approach has less error in all three cases and works well until $k$ equal to 15.

### B. Multi-scale deformation graph

Next, we evaluate the advantages of our multi-scale deformation graph compared to single deformation graph-based approaches. We are using two graphs: dense graph $G_d$ with a sampling density of 10mm and sparse graph $G_s$ with a sampling density of 25mm. State-of-the-art

---

[1]MakeHuman http://www.makehumancommunity.org
[2]Blender https://www.blender.org
[3]Motionbuilder-friendly BVH conversion CMU's Motion Capture Database

[4]Our Synthetic data is publicly available at https://research.csiro.au/robotics/databases or https://research.qut.edu.au/saivt/

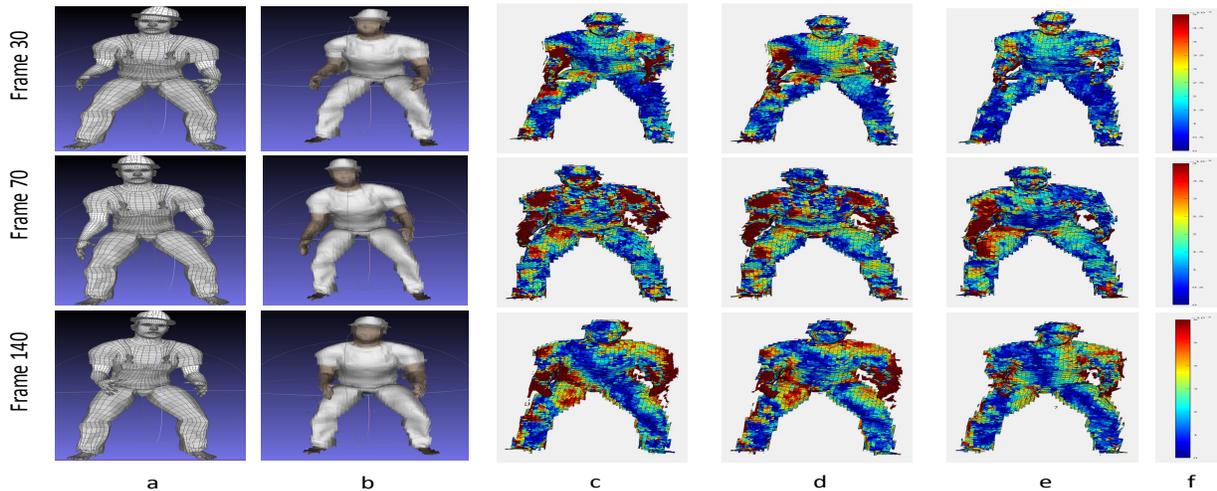

Fig. 6: Comparison of reconstruction using our multiscale deformation graph and single deformation graph. Each row belong to specific frame instants (30,70 and 140) a) ground truth geometry b) live reconstruction from our method. c,d,e - reconstruction using 25mm 30mm graphs and our method, which are color-coded based on the distance from the ground truth.

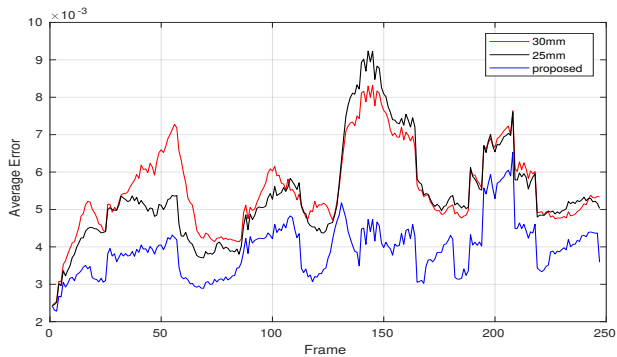

Fig. 7: Average error plot of our multi-scale approach against single deformation graph approaches. Average error is plotted agaitns each frame index. Red and black corresponds to reconstruction using 30mm, 25mm deformation graphs respectively and blue corresponds to our approach.

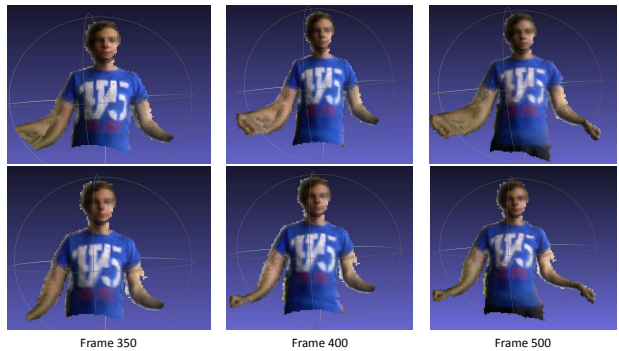

Fig. 8: An example where using only dense deformation graph (sampling density 10mm) fails in tracking. Upper row is the reconstruction using only dense graph and lower row is the reconstruction using our proposed multiscale approach.

|  | Case 1 | | Case 2 | | Case 3 | |
| --- | --- | --- | --- | --- | --- | --- |
|  | mean | std. | mean | std. | mean | std. |
| DynamicFusion | 7.5 | 2.4 | 67.2 | 45.3 | 64.7 | 36.9 |
| Our method | **7.2** | **2.3** | **9.6** | **6.6** | **7.0** | **3.2** |

TABLE II: Comparison of the squared distance error from ground truth by using DynamicFusion [4] and our method. The unit for error is mm.

approaches [4], [6] uses single graph of sampling density of 25mm.

In order to evaluate the importance of our approach, we compare with single graph approaches of sampling density 25mm and 30mm respectively. Reconstruction in these three cases at different frame instances are shown in figure 6. The first and second columns show snapshots of ground truth and our reconstructions. Note that reconstructions are partial because these data sequences do not complete 360°loops. Third and fourth columns show the color-coded meshes from 30mm, 25mm graph, and our approach. The meshes are color-coded based on the shortest projective distance from the ground truth. We can visually see that our method always has less distance from the ground truth. Average error from each live reconstruction is plotted in Figure 7. We observe that multi-scale approach always has a smaller average error (3.9mm average over full sequence) compared with single deformation graph (5.6mm and 5.4mm respectively) approaches.

Next, we experimentally show the importance of sparse graph $G_S$. For that, we reconstruct only using the dense graph and compared with our method. Figure 8 shows the visual comparison of reconstruction in both cases. We can see that even-though dense deformation graph can increase the overall reconstruction quality, it fails to track the motion of the hand. Note that state-of-the-art approaches [4], [6], use only one deformation graph. Therefore they are either sacrificing reconstruction quality or tracking quality. Our multi-scale deformation graph approach overcomes this problem and is the efficient solution to improve the tracking quality without compromising reconstruction quality.

Note that VolumeDeform [5] also uses multiscale strategy for optimization but because of they uses volumetric graph, they have to limit the finer graph's resolution to 45 to 50mm for a TSDF volume of 3m$^3$ to make the algorithm suitable for real-time operation. Since our approach is acting on the direct scene graph we are able to make finer graph up-to 10mm for modeling deformations.

## VI. Conclusion

We presented a novel framework for non-rigid 3D reconstruction that uses camera pose estimated from the background to achieve more robustness in handling the larger frame-to-frame motions. Our multiscale deformation graph enables tracking a wider range of deformations without compromising the quality of reconstruction. We also release a synthetic dataset for quantitative evaluation of RGB-D based non-rigid reconstruction methods. This dataset provides per frame ground truth geometry of the scene and trajectory of the camera.